\documentclass[11pt,a4paper]{article}
\usepackage[nohyperref]{acl2017}
\usepackage{times}
\usepackage{latexsym}
\usepackage{enumitem}
\usepackage{url}
\usepackage{graphicx}
\usepackage[colorinlistoftodos]{todonotes}
\aclfinalcopy 

\newcommand{\comment}[1]{}

\title{Predicting Audience's Laughter Using Convolutional Neural Network}

\author{Lei Chen\\
  Educational Testing Service (ETS)\\
  Princeton, NJ USA\\
  {\tt LChen@ets.org} \\\And
  Chong Min Lee \\
  Educational Testing Service (ETS) \\
  Princeton, NJ USA \\
  {\tt CLee001@ets.org} \\}

\date{}

\begin{document}
\maketitle
\begin{abstract}
For the purpose of automatically evaluating speakers' humor usage, we build a presentation corpus containing humorous utterances based on TED talks. Compared to previous data resources supporting humor recognition research, ours has several advantages, including (a) both positive and negative instances coming from a homogeneous data set, (b) containing a large number of speakers, and (c) being open. Focusing on using lexical cues for humor recognition, we systematically compare a newly emerging text classification method based on Convolutional Neural Networks (CNNs) with a well-established conventional method using linguistic knowledge. The advantages of the CNN method are both getting higher detection accuracies and being able to learn essential features automatically. 
\end{abstract}

\section{Introduction}
\label{sec:intro}

%
The ability to make effective presentations has been found to be linked with success at school and in the workplace.
Humor plays an important role in successful public speaking, e.g., helping to reduce public speaking anxiety often regarded as the most prevalent type of social phobia, generating shared amusement to boost persuasive power, and serving as a means to attract attention and reduce tension~\cite{Xu2016}.

Automatically simulating an audience's reactions to humor will not only be useful for presentation training, but also improve conversational systems by giving machines more empathetic power.
The present study reports our efforts in recognizing utterances that cause laughter in presentations.  These include building a corpus from TED talks and using Convolutional Neural Networks (CNNs) in the recognition.

The remainder of the paper is organized as follows: Section~\ref{sec:prev} briefly reviews the previous related research; Section~\ref{sec:data} describes the corpus we collected from TED talks; Section~\ref{sec:method} describes the text classification methods; Section~\ref{sec:exp} reports on our experiments; finally, Section~\ref{sec:diss} discusses the findings of our study and plans for future work.

\section{Previous Research}
\label{sec:prev}

Humor recognition refers to the task of deciding whether a sentence/spoken-utterance expresses a certain degree of humor. In most of the previous studies~\cite{mihalcea-strapparava:2005:HLTEMNLP,Purandare2006HumorPA,yang-EtAl:2015:EMNLP2}, humor recognition was modeled as a binary classification task.

%
In the seminal work \cite{mihalcea-strapparava:2005:HLTEMNLP}, a corpus of $16{,}000$ ``one-liners" was created using daily joke websites to collect humorous instances while using formal writing resources (e.g., news titles) to obtain non-humorous instances. 
Three humor-specific stylistic features, including \emph{alliteration}, \emph{antonymy}, and \emph{adult slang} were utilized together with content-based features to build classifiers.
In a recent work~\cite{yang-EtAl:2015:EMNLP2}, 
a new corpus was constructed from the \emph{Pun of the Day} website. \citet{yang-EtAl:2015:EMNLP2} explained and computed latent semantic structure features based on the following four aspects: (a) Incongruity, (b) Ambiguity, (c) Interpersonal Effect, and (d) Phonetic Style. In addition, Word2Vec~\cite{Mikolov2013} distributed representations were utilized in the model building. 
 
Beyond lexical cues from text inputs, other research has also utilized speakers' acoustic cues~\cite{Purandare2006HumorPA,Bertero2016_LREC}. These studies have typically used audio tracks from TV shows and their corresponding captions in order to categorize characters' speaking turns as humorous or non-humorous. Utterances prior to canned laughter that was manually inserted into the shows were treated as humorous, while other utterances were treated as negative cases.
%

Convolutional Neural Networks (CNNs) have recently been successfully used in several text categorization
tasks (e.g., review rating, sentiment recognition, and question type recognition). \newcite{Kim2014,Johnson2015,Zhang2015} suggested that using a simple CNN setup, which entails one layer of convolution on top of word embedding vectors, achieves excellent results on multiple tasks. 
%
%
Deep learning recently has been applied to computational humor research \cite{Bertero2016_LREC,Bertero2016_NAACL}. 
In \newcite{Bertero2016_LREC}, CNN was found to be the best model that uses both acoustic and lexical cues for humor recognition.
By using Long Short Time Memory (LSTM) cells~\cite{Hochreiter1997}, \newcite{Bertero2016_NAACL} showed that Recurrent Neural Networks (RNNs) perform better on modeling sequential information than Conditional Random Fields (CRFs)~\cite{Lafferty2001}. 

From the brief review, it is clear that corpora used in humor research so far are limited to one-line puns or jokes and conversations from TV comedy shows. There is a great need for an open corpus that can support investigating humor in presentations.\footnote{While we were working on this paper, we found a recent Master's thesis~\cite{acosta:2016:thesis} that also conducted research on detecting laughter on the TED transcriptions. However, that study only explored conventional text classification approaches.
}
CNN-based text categorization methods have been applied to humor recognition (e.g., in \cite{Bertero2016_LREC}) but with limitations: (a) a rigorous comparison with the state-of-the-art conventional method examined in \newcite{yang-EtAl:2015:EMNLP2} is missing; (b) CNN's performance in the previous research is not quite clear\footnote{Though CNN works best when using both lexical and acoustic cues, it did not outperform the Logistical Regression (LR) model when using text inputs exclusively.}; and (c) some important techniques that can improve CNN performance (e.g., using varied-sized filters and dropout regularization~\cite{Hinton2012}) were not applied. Therefore, the present study is meant to address these limitations. 

\section{TED Talk Data}
\label{sec:data}

TED Talks\footnote{\url{http://www.ted.com}} are recordings from TED conferences and other special TED programs. In the present study, we focused on the transcripts of the talks. Most transcripts of the talks contain the markup `(Laughter)', which represents where audiences laughed aloud during the talks. This special markup was used to determine utterance labels.

\begin{figure*}
\begin{description}[noitemsep]
\item[sent-7] \ldots 
\item[\ldots] \ldots
\item[No-Laughter] He has no memory of the past, no knowledge of the future, and he only cares about two things: easy and fun.
\item[sent-1] Now, in the animal world, that works fine. 
\item[Laughter] \emph{If you're a dog and you spend your whole life doing nothing other than easy and fun things, you're a huge success!} (Laughter)
\item[sent+1] And to the Monkey, humans are just another animal species.
\item[\ldots] \ldots
\item[sent+7] \ldots
\end{description}
\caption{An excerpt from TED talk \emph{``Tim Urban: Inside the mind of a master procrastinator"}
(\url{http://bit.ly/2l1P3RJ})}
\label{fig:excerpt}
\end{figure*}

We collected $1{,}192$ TED Talk transcripts\footnote{The transcripts were collected on 7/9/2015.}.
An example transcription is given in Figure~\ref{fig:excerpt}.
%
The collected transcripts were split into sentences using the Stanford CoreNLP
tool~\cite{manning-EtAl:2014:P14-5}. In this study, sentences containing or
immediately followed by `(Laughter)' were used as `Laughter' sentences, as shown in
Figure~\ref{fig:excerpt}; all other sentences were defined as `No-Laughter'
sentences. Following \citet{mihalcea-strapparava:2005:HLTEMNLP} and \citet{yang-EtAl:2015:EMNLP2}, we selected the same numbers ($n=4726$) of `Laughter' and `No-Laughter' sentences. To minimize possible topic shifts between positive and negative instances, for each positive instance, we picked one negative instance nearby (the context window was $7$ sentences in this study).
For example, in Figure~\ref{fig:excerpt}, a negative instance (corresponding to `sent-2')  was selected from the nearby sentences ranging from `sent-7' to `sent+7'.

\section{Methods}
\label{sec:method}
\vspace{-0.1in}

\subsection{Conventional Model}

Following \newcite{yang-EtAl:2015:EMNLP2}, we applied Random Forest~\cite{breiman_random_2001} to perform humor recognition by using the following two groups of features.
The first group are latent semantic structural features covering the following $4$ categories\footnote{The number in parenthesis indicates how many features are in that category}: 
\emph{Incongruity} ($2$), \emph{Ambiguity} ($6$), \emph{Interpersonal Effect} ($4$), and \emph{Phonetic Pattern} ($4$).  
The second group are semantic distance features, including the humor label classes from $5$ sentences in the training set that are closest to this sentence (found by using a \emph{k}-Nearest Neighbors (kNN) method), and each sentence's averaged Word2Vec representations ($n=300$). More details can be found in \citet{yang-EtAl:2015:EMNLP2}.

\subsection{CNN model}
\label{sec:model_CNN}

\begin{figure*} 
  \includegraphics[width=\textwidth]{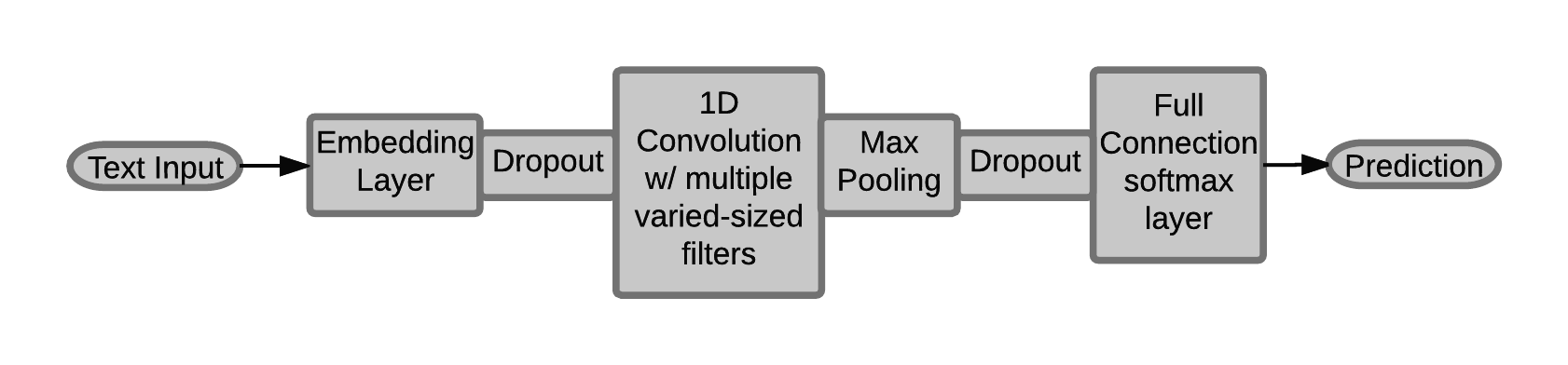}
  \caption{CNN network architecture }
  \label{fig:CNN}
\end{figure*}

Our CNN-based text classification's setup follows \newcite{Kim2014}. Figure~\ref{fig:CNN} depicts the model's details.
From the left side's input texts to the right side's prediction labels, different shapes of tensors flow through the entire network for solving the classification task in an end-to-end mode. 

Firstly, tokenized text strings were converted to a $2D$ tensor with shape $(L\times d)$, where $L$ represents sentences' maximum length while $d$ represents the word-embedding dimension. In this study, we utilized the Word2Vec~\cite{Mikolov2013} embedding vectors ($d=300$) that were trained on $100$ billion words of Google News. Next, the embedding matrix was fed into a $1D$ convolution network with multiple filters. To cover varied reception fields, we used filters of sizes of $f_{w}-1$, $f_{w}$, and $f_{w}+1$. For each filter size, $n_{f}$ filters were utilized. Then, max pooling, which stands for finding the largest value from a vector, was applied to each feature map (total $3\times n_{f}$ feature maps) output by the $1D$ convolution. Finally, maximum values from all of $3\times n_{f}$ filters were formed as a flattened vector to go through a fully connected (FC) layer to predict two possible labels (Laughter vs. No-Laughter). Note that for $1D$ convolution and FC layer's input, we applied `dropout'~\cite{Hinton2012} regularization, which entails randomly setting a proportion of network weights to be zero during model training, to overcome over-fitting. By using cross-entropy as the learning metric, the whole sequential network (all weights and bias) could be optimized by using any SGD optimization, e.g., Adam~\cite{adam}, Adadelta~\cite{adadelta}, and so on.

\section{Experiments}
\label{sec:exp}
\vspace{-0.1in}

We used two corpora: the TED Talk corpus (denoted as TED) and the Pun
of the Day corpus\footnote{The authors of \newcite{yang-EtAl:2015:EMNLP2} kindly
shared their data with us. We would like to thank them for their generosity.} (denoted as Pun). Note that we normalized words in the Pun data to lowercase to avoid a possibly elevated result caused by a special pattern: in the original format, all negative instances started with capital letters.  The Pun data allows us to verify that our implementation is consistent with the work reported in \newcite{yang-EtAl:2015:EMNLP2}.


In our experiment, we firstly divided each corpus into two parts. The smaller part (the Dev set) was used for setting various hyper-parameters used in text classifiers. The larger portion (the CV set) was then formulated as a $10$-fold cross-validation setup for obtaining a stable and comprehensive model evaluation result.
For the PUN data, the Dev contains $482$ sentences, while the CV set contains $4344$ sentences. For the TED data, the Dev set contains $1046$ utterances, while the CV set contains $8406$ utterances.
Note that, with a goal of building a speaker-independent humor detector, when partitioning our TED data set, we always kept all utterances of a single talk within the same partition. 
To our knowledge, this is the first time that such a strict experimental setup has been used in recognizing humor in conversations, and it makes the humor recognition task on the TED data quite challenging.  



When building conventional models, we developed our own feature extraction scripts and used 
the SKLL\footnote{\url{https://github.com/EducationalTestingService/skll}} python package for building  Random Forest models.
%
When implementing CNN, we used the Keras\footnote{https://github.com/fchollet/keras} Python package.\footnote{The implementation will be released with the paper.}
%
Regarding hyper-parameter tweaking, we utilized the Tree Parzen Estimation (TPE) method as detailed in \newcite{TPE}. 
%
%
After running $200$ iterations of tweaking, we ended up with the following selection: $f_{w}$ is $6$ (entailing that the various filter sizes are $(5,6,7)$), $n_{f}$ is $100$, $dropout_{1}$ is $0.7$ and $dropout_{2}$ is $0.35$, optimization uses Adam~\cite{adam}. 
When training the CNN model, we randomly selected $10\%$ of the training data as the validation set for using early stopping to avoid over-fitting.

\begin{table}
\begin{center}
\begin{tabular}{| c || r | r | r | r |}
\hline
           & Acc. (\%) & F1 & Precision & Recall \\
  \hline \hline
  \multicolumn{5}{|c|}{Pun set}\\
  \hline
  Chance& $50.2$       &   $.498$    &    $.506$    &   $.497$\\
  Base  & $78.3$       &   $.795$    &    $.757$    &   $.839$\\
  CNN   & \bf{86.1}    &   \bf{.857} &    \bf{.864}  &   \bf{.864}\\
  \hline
  \multicolumn{5}{|c|}{TED set}\\
  \hline
  Chance& $51.0$       &   $.506$    &    $.510$    &   $.503$\\
  Base  & $52.0$       &   $.595$    &    $.515$    &   $.705$\\
  CNN   & \bf{58.9}    &   \bf{.606} &    \bf{.582} &   $.632$\\
\hline
\end{tabular}
\end{center}
\caption{Humor recognition on both Pun and TED data sets by using (a) random prediction (Chance), conventional method (Base) and CNN method}
\label{tab:perf_pun_on_talk}
\end{table}

On the Pun data, the CNN model shows consistent improved performance over the conventional model, as suggested in \citet{yang-EtAl:2015:EMNLP2}. In particular, precision has been greatly increased from $0.762$ to $0.864$. On the TED data, we also observed that the CNN model helps to increase precision (from $0.515$ to $0.582$) and accuracy (from $52.0\%$ to $58.9\%$). The empirical evaluation results suggest that the CNN-based model has an advantage on the humor recognition task.
In addition, focusing on the system development time, generating and implementing those features in the conventional model would take days or even weeks. However, the CNN model automatically learns its optimal feature representation and can adjust the features automatically across data sets. This makes the CNN model quite versatile for supporting different tasks and data domains.
Compared with the humor recognition results on the Pun data, the results on the TED data are still quite low, and more research is needed to fully handle humor in authentic presentations.

\section{Discussion}
\label{sec:diss}
\vspace{-0.1in}

For the purpose of monitoring how well speakers can use humor during their presentations, we have created a corpus from TED talks.
Compared to the existing (albeit limited) corpora for humor recognition research,  ours  has the following advantages: (a) it was collected from authentic talks, rather than from TV shows performed by professional actors based on scripts; (b) it contains about $100$ times more speakers compared to the limited number of actors in existing corpora.  
We compared two types of leading text-based humor recognition methods: a conventional classifier (e.g., Random Forest) based on human-engineered features vs. an end-to-end CNN method, which relies on its inherent representation learning. We found that the CNN method has better performance. More importantly, the representation learning of the CNN method makes it very efficient when facing new data sets. 

Stemming from the present study, we envision that more research is worth pursuing: (a) for presentations, cues from other modalities such as audio or video will be included, similar to \newcite{Bertero2016_LREC}; (b) context information from multiple utterances will be modeled by using sequential modeling methods.

\bibliography{cnn_humor}
\bibliographystyle{acl_natbib}

\end{document}